\pdfoutput=1

\documentclass[11pt]{article}

\usepackage[preprint]{acl}

\usepackage{times}
\usepackage{latexsym}
\usepackage{booktabs}
\usepackage{amssymb}
\usepackage{amsmath}
\usepackage{subcaption}
\usepackage{comment}
\usepackage[T1]{fontenc}

\usepackage[utf8]{inputenc}

\usepackage{microtype}

\usepackage{inconsolata}

\usepackage{graphicx}

\title{CSIRO-LT at SemEval-2025 Task 11: Adapting LLMs for Emotion Recognition for Multiple Languages}

\author{
Jiyu Chen$^{1*}$, Necva B{\"o}l{\"u}c{\"u}$^{1*}$, Sarvnaz Karimi$^1$, Diego Moll\'a$^{2,1}$, C{\'e}cile Paris$^{1,2}$
\\
\begin{tabular}{cccc}
\multicolumn{2}{c}{$^{1}$CSIRO Data61, Australia}&
\multicolumn{2}{c}{$^{2}$Macquarie University, Australia} \\
\multicolumn{2}{c}{\tt firstname.lastname@csiro.au}&
\multicolumn{2}{c}{\tt diego.molla-aliod@mq.edu.au}\\
\multicolumn{4}{c}{\small $^*$ indicates co-first authors.}
\end{tabular}
}

\begin{document}
\maketitle
\begin{abstract}
Detecting emotions across different languages is challenging due to the varied and culturally nuanced ways of emotional expressions. The \textit{Semeval 2025 Task 11: Bridging the Gap in Text-Based emotion} shared task was organised to investigate emotion recognition across different languages. The goal of the task is to implement an emotion recogniser that can identify the basic emotional states that general third-party observers would attribute to an author based on their written text snippet, along with the intensity of those emotions. We report our investigation of various task-adaptation strategies for LLMs in emotion recognition. We show that the most effective method for this task is to fine-tune a pre-trained multilingual LLM with LoRA setting separately for each language.
\end{abstract}

\section{Introduction}

Text-based emotion recognition plays a crucial role in studies related to mental health~\citep{golder2011diurnal}, emotional intelligence~\citep{turcan2021emotion}, and human-computer interaction~\citep{li2022knowledge}. However, recognising emotions in text remains a significant challenge, especially across different languages~\citep{mohammad2018semeval}. Linguistic variations, cultural differences in emotional expression, and the scarcity of annotated data for low-resourced languages make emotion recognition particularly complex~\citep{barrett2011context,lindquist2013s,schroder2013affective}. 

The \textit{SemEval-2025 Task 11: Bridging the Gap in Text-Based Emotion} shared task~\citep{muhammad-etal-2025-semeval}\footnote{\url{https://github.com/emotion-analysis-project/SemEval2025-task11}} is part of the International Workshop on Semantic Evaluation (SemEval). Its objective is to detect {\em collectively perceived emotions} within text snippets written in 32 different languages. Collectively perceived emotion refers to the basic emotional states---such as anger, joy, and disgust---that third-party observers from the general public can attribute to a text snippet generated by a writer. Unlike tasks aimed at identifying the writer's actual emotional states or the emotional states evoked in individual reader~\citep{mohammad2022ethics,mohammad2023best}, this task emphasises the shared perception of emotions by general readers. This distinction is crucial, as perceived emotions can vary significantly from both intended and personally experienced emotions.

The shared task consists of three tracks: (A) \textit{Multi-label Emotion Detection}, where the goal is to predict the perceived emotional states expressed in a given text snippet, including joy, sadness, fear, anger, surprise, and disgust; (B) \textit{Emotion Intensity},  where the objective is to predict the intensity scale of each perceived emotional state for a given text snippet. Each emotional state is rated on a 4-point categorical scale: 0 (no emotion), 1 (low intensity), 2 (moderate intensity), and 3 (high intensity); and~(C) \textit{Cross-lingual Emotion Detection}, where the goal is to predict the perceived emotion on text snippets written in a language different from the one used for model training, such as training on English-written data but making emotion recognition on Javanese-written data.

Our team participated in Tracks A and B, focusing on fine-tuning Large Language Models (LLMs). LLMs have contributed to significant improvements across different NLP tasks~\citep{brown2020language,touvron2023llama}. However, previous studies suggest that they are ineffective for emotion classification in zero-shot and few-shot settings, even when provided with In-Context Learning (ICL) prompts~\citep{liu2024emollms}. A potential improvement can be achieved by fine-tuning LLMs with instructions~\citep{zhang2023instruction,liu2024emollms}. Drawing inspiration from~\citet{liu2024emollms}, we explore instruction-tuning and continual fine-tuning of multilingual LLMs. We then compare the effectiveness of this adaptation for emotion recognition across different languages by fine-tuning a multilingual model individually for each language. We also propose several adaptation approaches and conduct a comparative analysis across these approaches, specifically for Track A.
We develop an adaptation strategy for LLMs in emotion recognition that is effective across languages of varying resourced levels (see the categorisation of resource abundance by~\citet{joshi2020state,ustun2024aya}).

\section{Methods}\label{sec:method}
\subsection{Track A: Multi-label Emotion Detection}
We formulate this task as a \textbf{binary classification} problem for the detection of each emotional state. For each input, the LLMs predict the occurrence of the six emotions---anger, disgust, fear, joy, sadness, and surprise---independently, determining whether a specific emotional state is {\em present} (1) or {\em absent} (0). The final results are aggregated for the input, to represent a multi-label emotion recognition setting. To address class imbalance, we apply oversampling to balance the binary classification instances.

The proposed adaptation strategies are described below:
\begin{itemize}
    \item \textbf{Few-shot}: We apply ICL based on the BM25~\cite{bm25} scores to retrieve instances from the training set to construct a prompt for the prediction of each test instance (see the prompt template in Section~\ref{sec:hyperparams}). That is, we use BM25 scores to rank semantic (bag-of-words) relevancy between the given test instance (as a query) and each training instance (as a document)~\citep{schutze2008introduction, robertson2009probabilistic}. We retrieve the top $k$ most relevant instances and their manually annotated emotional state label as $k$-shot examples to prompt-tune LLMs for each test instance.

    \item \textbf{Supervised Fine-Tuning (SFT):} We fine-tune pre-trained and instruction-tuned multilingual LLMs using supervised and parameter-efficient settings.

    \item \textbf{English-bridged Adaptation (E-Bridge):} We apply SFT to LLMs on English-written instances and then apply continual SFT for the adaptation to other languages.
  
    \item \textbf{Marginalisation:} We first apply SFT to LLM on Track B (see details of SFT in Section~\ref{sec:track_b_adapt}), and then use fine-tuned LLM to make predictions for instances on Track A. To align with the binary classification in Track A, predictions of $1$–$3$ are marginalised into $1$, indicating the presence of an emotional state, while predictions of zero are retained to represent its absence.
\end{itemize}

\subsection{Track B: Emotion Intensity Detection}\label{sec:track_b_adapt}

We frame this task as a \textbf{multi-class classification} problem for each emotion. Given an input text, the LLM predicts whether a specific emotion can be perceived at a given intensity level. The intensity is measured on a four-point categorical scale: 0 for no emotion, 1 for low intensity, 2 for moderate intensity, and 3 for high intensity. Thus, for a text with six emotional states, the LLM processes the input six times, once for each emotion. The final results are aggregated to the input text for the completion of multi-label emotion intensity recognition set by Track B.

To achieve this, we convert each text with multiple emotion intensities into separate instances, one for each emotion. We fine-tune pre-trained and instruction-tuned LLMs on this transformed dataset, using supervised learning to predict the emotional intensity for each emotion independently (SFT) and apply zero-shot, where we only use instruction as a baseline. 

\begin{table*}[ht!]
\centering
\resizebox{0.9\textwidth}{!}
{
    \begin{tabular}{l ccl c ccl}
    \toprule
         & \multicolumn{3}{c}{\textbf{Track A} ($F_1$)}\ && \multicolumn{3}{c}{\textbf{Track B} ($r$)} \\
         \cmidrule{2-4}\cmidrule{6-8}
         \textbf{Languages} & Baseline & Ours & Model && Baseline & Ours & Model\\
    \midrule
         Afrikaans (afr) & 37.14 & 31.61 & aya-101&& --- & --- & ---\\
         Algerian Arabic (arq) & 41.41  & 52.55 & aya-32&& 1.64 & 52.11 & aya-32\\
         Amharic (amh) & 63.83 & 58.81 & aya-32 && 50.79 & 15.40 & llama\\
         Chinese (chn) & 53.08 & 57.84 & aya-32 && 40.53 & 54.92 & aya-32\\
         Emakhuwa (vmw) & 12.14 & 17.27  & aya-101 && --- & ---& --- \\
         English (eng) & 70.83 & 77.62 & aya-32 && 64.15 & 72.09 & llama\\
         German (deu) & 64.23 & 65.72 & aya-32  && 56.21 & 60.98 & aya-32\\
         Hausa (hau) & 59.55 & 54.25  & aya-101 &&27.03 & 37.15 & llama\\
         Hindi (hin) & 85.51 & 73.16 & aya-32 && --- & ---& ---\\
         Igbo (ibo) & 47.90 & 32.56  & aya-101&& --- & ---& ---\\
         Indonesian (ind) & --- & --- & --- && ---& ---& ---\\
         isiXhosa (xho) & --- & --- & --- && ---& ---& ---\\
         isiZulu (zul) & --- & ---& --- && --- & ---& ---\\
         Javanese (jav) & --- & ---& --- && --- & ---& ---\\
         Kinyarwanda (kin) & 46.29 & 42.95  & aya-101 && --- & ---& ---\\
         Marathi (mar) & 82.22& 75.75  & aya-101 && --- & ---& ---\\
         Moroccan Arabic (ary) & 47.16 & --- & --- && --- & --- & ---\\         
         Nigerian-Pidgin (pcm) & 55.50 & 21.81 & aya-101 && --- & --- & ---\\
         Oromo (orm) & 12.63 & 39.24  & aya-101&& --- & ---& ---\\
         Portuguese (Brazil) (ptbr) & 42.57 & 55.54 & aya-32 && 29.74 & 48.88 & emo-aya\\
         Portuguese (Mozambican) (ptmz) &  45.91 & --- & --- && --- & --- & ---\\
         Romanian (ron) & 76.23 & 72.24 & aya-101 && 55.66 & 59.22 & aya-32\\
         Russian (rus) & 83.44 & 89.10 & aya-101 && 87.66 & 79.26 & llama\\
         Somali (som) & 45.93 & 43.28  & aya-101&& --- & --- & ---\\
         Spanish (Latin American) (esp) & 77.44 & 82.00 & aya-101 && 72.59 & 69.66 & llama\\
         Sundanese (sun) & 37.31  & 48.75 & aya-101 && --- & --- & ---\\
         Swahili (swa) & 22.65 & 30.31  & aya-101 && --- & --- & ---\\
         Swedish (swe) & 51.98 & 48.88  & aya-101 && --- & --- & ---\\
         Tatar (tat) & 53.94 & 53.84  & aya-101 && --- & --- & ---\\
         Tigrinya (tir) & 46.28 & 49.95  & aya-101 && --- & --- & ---\\
         Ukrainian (ukr) & 53.45 & 66.40 & aya-32 && 39.94 & 49.42 & emo-aya$^*$\\
         Yoruba (yor) & 9.22 & 29.96  & aya-101 && --- & --- & ---\\
         
    \bottomrule
    \end{tabular}
    }
    \caption{Effectiveness of baseline and our approaches (SFT or zero-shot (*) of a specific model) on the test set. The baseline results are provided by the task organisers~\citet{muhammad2025brighterbridginggaphumanannotated}. Note that aya-32 denotes the \texttt{aya-32b-expanse} model, llama denotes the \texttt{Llama3.1-8B-Instruct} model, emo-aya denotes instruction-tuned \texttt{aya-32b-expanse}.}
    \label{tab:test_results}
\end{table*}

\section{Experimental Setup}\label{sec:experimental_setup}
\subsection{Shared Task Dataset}

The dataset used in the shared task is a combination of EthioEmo~\citep{belay-etal-2025-evaluating} and BRIGHTER~\citep{muhammad2025brighterbridginggaphumanannotated}, which includes emotion annotations for multiple languages. Specifically, $28$ languages for Track A and $11$ languages for Track B. The statistical details of the annotated languages are detailed by~\citet{belay-etal-2025-evaluating} for Amharic, Oromo, Somali, and Tigrinya, and by~\citet{muhammad2025brighterbridginggaphumanannotated} for the remaining languages.


\paragraph{Instruction-tuning on External Dataset:} We leveraged an external dataset to instruction-tune LLMs~\citep{liu2024emollms}. This helps the model generalise on the external task by learning how to follow instructions for emotion recognition before fine-tuning on the dataset specific to this shared task.
The selected external dataset is an extension to the \textit{SemEval-2018 Task 1: Affect in Tweets}, which includes a series of subtasks related to affectual state inference: (1) emotion intensity
regression; (2) emotion intensity ordinal classification; (3) valence (sentiment) regression; (4) valence ordinal classification; and, (5) emotion
classification~\citep{mohammad-kiritchenko-2018-understanding, mohammad2018semeval} which overlaps with the tracks of this shared task.

\subsection{Evaluation Metrics}\label{ssec:evaliation_metric}
For both Track A and Track B, we follow the evaluation metrics specified by the shared task organisers. Track A uses the unweighted average of all per-emotion $F_1$ scores. Track B uses the average of per-emotion Pearson’s correlation coefficient ($r$).

\begin{table*}[tb]
    \centering
    \resizebox{\linewidth}{!}{
        \begin{tabular}{lccclcccclc}
            \toprule
             & \multicolumn{3}{c}{\bf Supervised Fine-tuning (SFT)} &&\multicolumn{4}{c}{\bf Marginalisation} && \bf{Few-shots}\\
             \cmidrule{2-4}\cmidrule{6-9}\cmidrule{11-11}
            \bf Language & AYA$_{ft}$ & EMO-AYA$_{ft}$ & E-Bridge && AYA$_{pt_b}$ & EMO-AYA$_{pt_b}$ & AYA$_{ft_b}$ & EMO-AYA$_{ft_b}$&&  \texttt{1-shot}\\
            
            \midrule
            English &\textbf{80.12}&73.31&---&& 64.55& 57.94 & 68.47 &72.88 && 52.31\\
            German &\textbf{62.31}&52.43&56.21&& 53.92& 46.65& 60.69 &54.56&&  30.24\\
            Portuguese &\textbf{61.29}&53.42&57.33&& 44.55& 40.56 & 41.87 &45.93 &&  34.11\\
            Russian &\textbf{89.11}&87.13&86.16&& 54.12& 60.56 & 72.42 &55.34&&  75.01\\
            \bottomrule
        \end{tabular}
        }
        \caption{Macro $F_1$ scores on the development split of Track A for four languages. AYA${ft}$ denotes direct SFT of the \texttt{aya-32b-expanse} model on the training set. EMO-AYA${ft}$ indicates first performing instruction-tuning on the \texttt{aya} model on the external dataset, followed by SFT on the training set. E-Bridge refers to SFT of the \texttt{aya} model on the training set in English, followed by continual SFT on the remaining three languages. AYA${pt_b}$ involves prompt-tuning a model in a zero-shot setting to complete the Track B objective, followed by marginalisation. EMO-AYA${pt_b}$ represents applying instruction-tuning to \texttt{aya} model on the external dataset and then prompt-tuning it in a zero-shot setting on Track B, followed by marginalisation. The best results are \textbf{boldfaced}.}
        \label{tab:results_a}
\end{table*}

\subsection{Hyperparameters}\label{sec:hyperparams}
\paragraph{LLMs:} We use three open-source multilingual LLMs: AYA (\texttt{aya-101}~\citep{ustun2024aya}, \texttt{aya-32b-expanse}~\citep{dang2024ayaexpansecombiningresearch}), and LLAMA (\texttt{Llama3.1-8B-Instruct}~\citep{dubey2024llama}). \texttt{aya-101} (mT5-based) offers broad multilingual support with a smaller size, while \texttt{aya-32b-expanse} (GPT-style) provides larger capacity and similarly wide language coverage. \texttt{Llama3.1-8B-Instruct} (GPT-style) is smaller but supports only seven languages. Due to its broader coverage and capacity, \texttt{aya-32b-expanse} was preferred, with \texttt{aya-101} as a lightweight alternative. For languages jointly supported by both \texttt{aya-32b-expanse} and \texttt{Llama3.1-8B-Instruct}, such as English and German, model selection also accounts for differences in parameter size.

We employ Low-Rank Adaptation (LoRA)~\citep{hu2022lora} and apply $4$-bits quantisation~\citep{jacob2018quantization} to LLMs for parameter-efficient SFT. We set the LoRA rank and alpha parameters to $32$ and 64, respectively. The dropout ratio is set to $0.05$. We limit both the input source length and the target length to $512$. The training epoch size is $10$ and the batch size is $2$. The learning rate is set to $2e-5$ for Track A and $5e-5$ for Track B.

The formulations of instructions for zero-shot, few-shot (ICL), and SFT settings are as below:

\begin{itemize}
    \item \textbf{Track A:} ``\textit{You are detecting emotions on a statement written in \{language\}. Statement: \{text\}. Does this statement express \{emotion\}? Answer 1 for yes and 0 for no.}''
    \item \textbf{Track B:} ``\textit{Task: Categorize the tweet into an intensity level of the specified emotion E, representing the mental state of the tweeter. 0: no E can be inferred. 1: low amount of E can be inferred. 2: moderate amount of E can be inferred. 3: high amount of E can be inferred. Tweet: \{text\} Emotion \{emotion\} Intensity class:}''   
\end{itemize}
\noindent
where \textit{\{text\}} is the text content of each instance, and \textit{\{emotion\}} is one of the six emotional states (or five, excluding \texttt{disgust} for English and \texttt{Surprise} for Afrikaans). The instruction of Track B is adapted from~\citet{liu2024emollms}.

For BM25-based few-shot prompting, we use the \texttt{rank\_bm25} library~\citep{rankbm25} and choose the default parameter setting, $b=0.75$ and $k_1=1.5$, for BM25.

We use NVIDIA H100 GPUs running on one node for this experiment. 


\section{Experimental Results}\label{sec:result}

\subsection*{Results on Testing Set} 
We present the results of the submitted predictions for ranking (testing) in Table~\ref{tab:test_results}, including the baseline results per language provided by~\citet{muhammad2025brighterbridginggaphumanannotated}. We submitted results for $26$ out of 32 languages in Track A and the 11 (all) languages provided in Track B. 

We observed noticeable variations in effectiveness across languages for both Track A (macro $F_1$) and Track B (average $r$). The baseline approach is mostly effective in higher-resourced languages, such as German, English, and Russian. However, applying instruction-tuning or SFT to LLMs on mid-resourced and lower-resourced language is more effective than the baseline for emotion recognition. 

Additionally, model selection plays a crucial role, as larger GPT-based models like \texttt{aya-32b-expanse} outperform smaller mT5-based models like \texttt{aya-101}, particularly in lower-resourced languages where the latter struggles.

We applied the adaptation strategies for the test set prediction based on the highest $F_1$ and correlation coefficient $r$ achieved on the development set in Track A \& B, respectively. 
\begin{table*}[ht]
\resizebox{\textwidth}{!}
{
    \begin{tabular}{lcccclcccc}
    \toprule
    & \multicolumn{4}{c}{\bf Zero-shot} && \multicolumn{4}{c}{\bf SFT}\\\cmidrule{2-5}\cmidrule{7-10}
         \bf Language & AYA & EMO-AYA & LLAMA & EMO-LLAMA & & AYA$_{ft}$ & EMO-AYA$_{ft}$ & LLAMA$_{ft}$ & EMO-LLAMA$_{ft}$\\
    \midrule
         Algerian Arabic (arq)  & 41.55 & 46.39 & 16.65 & 37.44 & & \textbf{64.66} & 9.11 & 30.61 & 28.56\\
         Amharic (amh) & 7.63 & 7.37 & 11.93 & 16.40 & & --- & --- & \textbf{25.30} & 24.12 \\
         Chinese (chn) & 47.61 & 49.00 & 35.57 & 43.01 && \textbf{62.37} & 48.37 & 39.83 & 44.15\\
         English (eng) & 57.80 & 56.16 & 43.87 & 57.32 && 73.81 & 74.04 & \textbf{75.87} & 75.23\\
         German (deu) & 53.00 & 43.39 & 41.61 & 45.11 && \textbf{55.93} & 50.04 & 41.09 & 44.16\\
         Hausa (hau) & 15.08 & 16.77 & 16.15 & 15.34 && 21.31 & 16.18 & \textbf{45.13} & 42.56\\
         Portuguese (Brazilian) (ptbr) & 46.71 & 49.02 & 31.17 & 30.96 && 45.63 & \textbf{53.42} & --- & ---\\
         Romanian (ron) & 57.92 & 47.36 & 47.98 & 47.74 && \textbf{63.35} & 61.22 & 50.76 & 52.34\\
         Russian (rus) & 56.01 & 62.34 & 34.98 & 50.47 && 75.44 & 70.63 & \textbf{83.62} & 82.10\\
         Spanish (Latin American) (esp) & 57.77 & 54.77 & 45.80 & 48.69 && 68.50 & 64.36 & \textbf{70.07} & 69.10\\
         Ukrainian (ukr) & 45.68 & \textbf{50.07} & 24.12 & 33.05 && --- & ---& 41.99 & 40.56\\      
    \bottomrule
    \end{tabular}
    }
    \caption{The average $r$ on the development set of Track B. AYA and LLAMA refer to the base models, \texttt{aya-expanse-32b} and \texttt{Llama3.1-8B-Instruct}, respectively. EMO-AYA and EMO-LLAMA are their instruction-tuned  versions. *$_{ft}$ are fine-tuned versions. The best results are \textbf{boldfaced}.}
    \label{tab:results_b}
\end{table*}


\subsection*{Results on Track A Development Set}
We compared the effectiveness of the various adaptation strategies with experiments in English, German, Portuguese, and Russian. All comparisons are statistically validated using hypothesis testing with a significance threshold of
$p<0.05$.
We observed that directly applying SFT to LLMs on the training set (AYA$_{ft}$) consistently achieves the highest macro $F_1$ scores across all four languages, outperforming all of the other experimented adaptation approaches, such as: (i) BM25-based ICL ($1$-shot), (ii) instruction-tuning on the external dataset before SFT on the training set (EMO-AYA), (iii) bridging the adaptation with English (E-Bridge), or (iv) applying marginalisation to predictions of LLMs fine-tuned on Track B (Table~\ref{tab:results_a}).

These results suggest that LLMs may not significantly benefit from instruction tuning, as direct SFT shows greater effectiveness across all four languages. E-Bridge can be viewed as a specialised form of instruction tuning, where the LLM first learns instructions in English instances before being fine-tuned in other languages. This method proves effective for German and Portuguese but is less effective for Russian, possibly due to the closer cultural alignment of German and Portuguese speakers with English speakers~\citep{rinke2021portuguese,engdeurelation}.


\subsection*{Results on Track B Development Set}
The emotion intensity results across multiple languages using both base and instruction-tuned versions of AYA and LLAMA in zero-shot and SFT settings are shown in Table~\ref{tab:results_b}. Fine-tuning significantly improves performance across most languages, demonstrating that task-specific adaptation benefits intensity detection. While fine-tuning offers substantial gains across the languages, the benefits are often more pronounced in higher-resourced languages (i.e., English, Spanish). While AYA generally demonstrates stronger zero-shot performance, LLAMA benefits more from instruction tuning, showing significant improvements after fine-tuning. Instruction-tuned models (EMO-AYA and EMO-LLAMA) provide some advantages in zero-shot settings, particularly in languages with less training data, but their impact diminishes after fine-tuning, suggesting that instruction-tuning alone is often sufficient for intensity detection.

Higher-resourced languages, such as English, Russian, and Spanish, consistently achieve better results, with both zero-shot and fine-tuned models performing reliably. Mid-resourced languages, including German, Portuguese, and Romanian, show moderate performance, benefiting from fine-tuning but still exhibiting variability depending on the model. The performance of all languages improves with fine-tuning, but challenges persist in making significant gains for languages where the models initially perform poorly. Despite this, the advancements show that fine-tuning and instruction tuning can help optimise model behaviour across languages, and targeted adaptation strategies may further boost results for emotion intensity detection tasks.

\section{Conclusions}\label{sec:conclusion}
We participated in the \textit{SemEval 2025 Task 11: Bridging the Gap in Text-Based Emotion} shared task, which included tracks for multi-label emotion detection (Track A) and emotion intensity detection (Track B). For Track A, we approached it as a binary classification problem for each emotion---determining whether an emotional state is present in or absent from a given input text snippet. We found that direct supervised fine-tuning can effectively adapt LLMs for the detection of emotions for most languages, except for the lower-resourced languages where few-shot learning is more effective. For Track B, we achieved the best results by employing different LLMs (both direct SFT and instruction-tuned) for each language. The results in both tracks suggest that, when adapting LLMs for emotion recognition on most mid-resourced and higher-resourced language, instruction-tuning was not as effective as in other NLP tasks. A more suitable approach is to directly apply supervised fine-tuning of LLMs on task-specific datasets.

\section*{Limitations}
The exploration of various adaptation strategies was limited to four languages (English, German, Russian, and Portuguese), which may not generalise to other languages, particularly lower-resourced ones or those with different linguistic structures. The models used may reflect biases from the training data, which could affect performance in low-resourced languages. We only explored prompt-tuning and instruction-tuning with the parameter-efficient LoRA setting for adapting LLMs. 


\section*{Ethical Considerations}
We relied on the dataset providers to remove any material from the dataset that may reveal anyone's identity in their posts used in this study. We guarantee that datasets are only used for scientific or research purposes and are not redistributed or shared with third parties. This project is subject to the ethics approval and agreement provided by the SemEval-2025 task organisers.

\bibliography{custom}

\end{document}